\documentclass[10pt,twocolumn,letterpaper]{article}

\usepackage[pagenumbers]{cvpr} % To force page numbers, e.g. for an arXiv version

\usepackage[dvipsnames]{xcolor}

\usepackage{enumitem}
\usepackage{graphicx}
\usepackage{booktabs}
\usepackage{rotating}
\usepackage{array}
\usepackage{tabularx}
\usepackage{float}

\newif\ifcomments
\commentstrue

\newif\ifarxiv
\arxivfalse

\newcommand{\textToken}{\boldsymbol{c}}

\definecolor{cvprblue}{rgb}{0.21,0.49,0.74}
\usepackage[pagebackref,breaklinks,colorlinks,citecolor=cvprblue]{hyperref}

\def\paperID{13159} % *** Enter the Paper ID here
\def\confName{CVPR}
\def\confYear{2024}
\newcommand{\OURS}{OrientDream}

\title{\OURS: Streamlining Text-to-3D Generation with\\ Explicit Orientation Control}

\author{
    Yuzhong Huang\textsuperscript{1,2,$*$} \quad
    Zhong Li\textsuperscript{1} \quad
    Zhang Chen\textsuperscript{1} \quad
    Zhiyuan Ren \textsuperscript{1} \quad\\
    Guosheng Lin \textsuperscript{3} \quad
    Fred Morstatter \textsuperscript{2} \quad
    Yi Xu\textsuperscript{1}\\
    \\
    \textsuperscript{1} OPPO US Research Center \quad
    \textsuperscript{2} University of Southern California \\
    \textsuperscript{3} Nanyang Technological University
    \\
}

\begin{document}
\twocolumn[{%
	\renewcommand\twocolumn[1][]{#1}%
	\maketitle
	\begin{center}

        \vspace{-1em}
    \setlength\tabcolsep{1pt}
    \begin{tabularx}{1.0\linewidth}{cc}
    \includegraphics[width=0.5\linewidth]{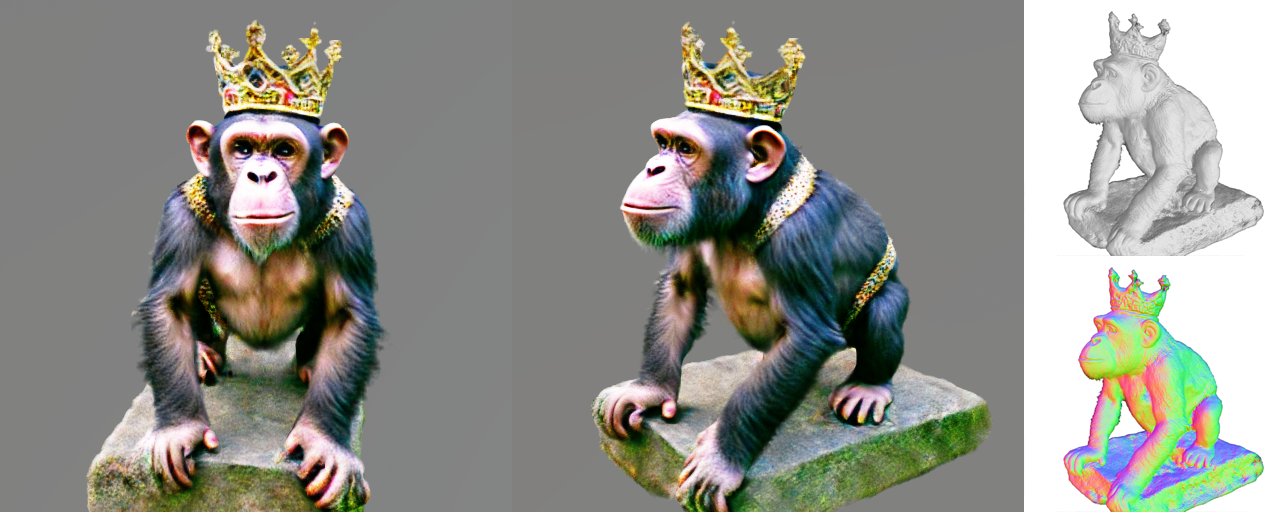} &
    \includegraphics[width=0.5\linewidth]{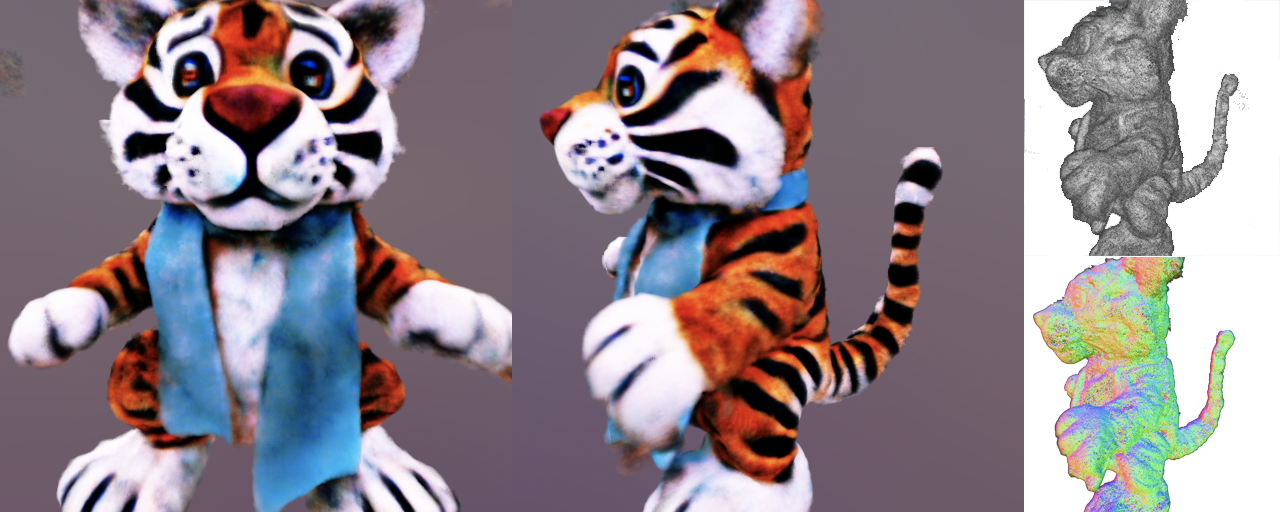}\\
     a chimpanzee dressed like Henry VIII king of England
    &  a small tiger dressed with sash  \\
    \includegraphics[width=0.5\linewidth]{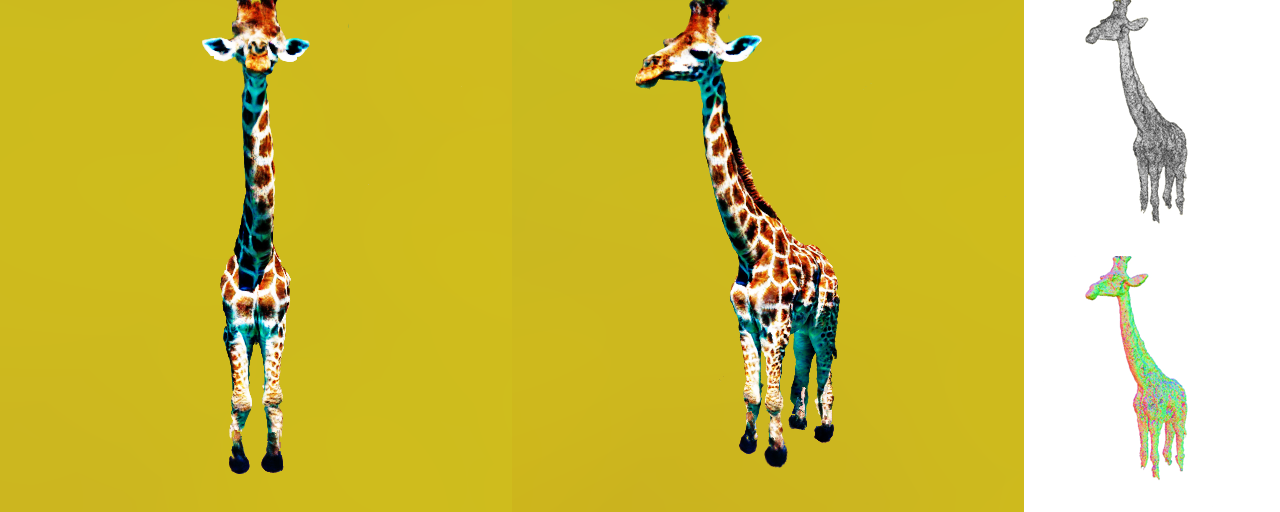} & \includegraphics[width=0.5\linewidth]{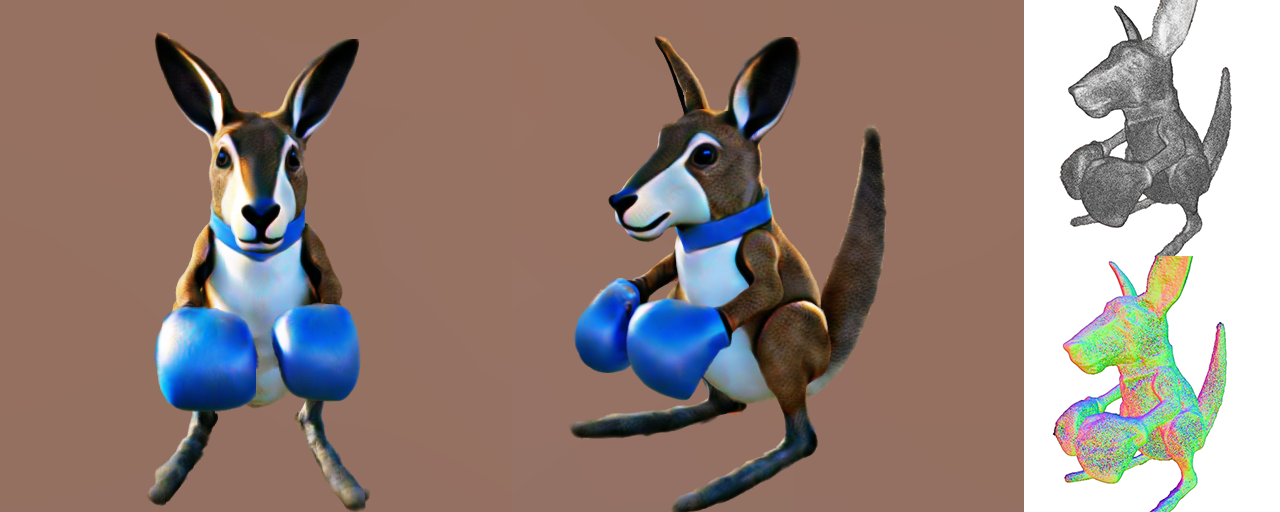}\\ 
    a majestic giraffe with a long neck & a kangaroo wearing boxing gloves \\
    \bottomrule
    \end{tabularx}
\end{center}
\vspace{-1em}
    \captionof{figure}{
                \textbf{3D objects generated by \OURS{} pipeline using text input. } Our \OURS{} pipeline successfully generates 3D objects with high-quality textures, effectively free from the view inconsistencies commonly referred to as the Janus problem. We also present visualizations of geometry and normal maps alongside each result for further detail and clarity.
		}
      \vspace{1em}
		\label{fig:teaser}
}]
\begin{abstract}

 {\let\thefootnote\relax\footnote{{
{$^{*}$} Work done while Yuzhong was an intern at OPPO US Research Center.
 }}}

% {\let\thefootnote\relax\footnote{{
% {$^{\dagger}$ Corresponding author}.
%  }}}

% Recent advances in text-to-3D modeling have combined text-to-image models with 3D representation like NeRF to generate high-quality 3D objects from text prompts. However, existing methods~\cite{poole2022dreamfusion} suffer from (1) Janus problem as the 2D guidance model is not aware of camera location. (2) slow learning speed as thousands of optimization steps are required. 

 % % First
% We propose \OURS, a pose-conditioned diffusion model that is able to generate photo realistic 2D images at given camera locations and text prompt.

% % Second
% We demonstrated that it can serve as a 3D consistent 2D prior. When using \OURS in text-to-3D modeling, it could provide view-dependent update gradient at the different camera locations, thus addressing the root cause of Janus problem.

% % Third
% Given the ability of \OURS to generate views at given location, we implemented decoupled back propagation, which perform multiple NeRF update per diffusion model evaluation, and greatly reduce the learning time.
% % ~\cite{liu2023zero, shi2023mvdream,stable-dreamfusion,lin2022magic3d,wang2023prolificdreamer}
% % Fourth
% Compared to existing methods~\cite{liu2023zero, shi2023mvdream} that learning on 3D dataset~\cite{objaverse}, our method is learned on pose annotated, real-world captured 2D dataset~\cite{yu2023mvimgnet}. which didn't suffer from reduced texture quality and limited variety. 

In the evolving landscape of text-to-3D technology, Dreamfusion~\cite{poole2022dreamfusion} has showcased its proficiency by utilizing Score Distillation Sampling (SDS) to optimize implicit representations such as NeRF. This process is achieved through the distillation of pretrained large-scale text-to-image diffusion models. However, Dreamfusion encounters fidelity and efficiency constraints: it faces the multi-head Janus issue and exhibits a relatively slow optimization process. To circumvent these challenges, we introduce \OURS, a camera orientation conditioned framework designed for efficient and multi-view consistent 3D generation from textual prompts. Our strategy emphasizes the implementation of an explicit camera orientation conditioned feature in the pre-training of a 2D text-to-image diffusion module. This feature effectively utilizes data from MVImgNet, an extensive external multi-view dataset, to refine and bolster its functionality. Subsequently, we utilize the pre-conditioned 2D images as a basis for optimizing a randomly initialized implicit representation (NeRF). This process is significantly expedited by a decoupled back-propagation technique, allowing for multiple updates of implicit parameters per optimization cycle. Our experiments reveal that our method not only produces high-quality NeRF models with consistent multi-view properties but also achieves an optimization speed significantly greater than existing methods, as quantified by comparative metrics.

% resolution,speed

\end{abstract}
    
\section{Introduction}
\label{sec:intro}

The demand for 3D digital content creation has surged recently, driven by advancements in user-end platforms ranging from industrial computing to personal, mobile, and virtual metaverse environments, impacting sectors like retail and education. Despite this growth, 3D content creation remains challenging and often requires expert skills. Integrating natural language into 3D content creation could democratize this process. Recent developments in text-to-2D image creation techniques, especially in diffusion models~\cite{rombach2022high}, have shown significant progress, supported by extensive online 2D image data. However, text-to-3D generation faces challenges, notably the lack of comprehensive  3D training data. Although quality 3D datasets exist~\cite{bai2023ffhq,choi2020stargan}, they are limited in scale and diversity. Current research, such as in Lin et al.~\cite{lin2022magic3d}, focusing on specific 3D object categories like human faces, lacks the versatility needed for broader artistic applications.

Recent advancements, particularly in DreamFusion~\cite{xu2023dream3d}, have showcased that pre-trained 2D diffusion models, which utilize 2D image data, can effectively serve as a 2D prior in the optimization of 3D Neural Radiance Fields (NeRF) representations through Score Distillation Sampling (SDS). This approach aligns NeRF-rendered images with photorealistic images derived from text prompts, facilitating a novel method for text-to-3D generation that bypasses the need for actual 3D content data. However, this technique faces notable challenges. The lack of relative pose information in the 2D prior supervision leads to the "multi-head Janus issue," as evident in Fig.~\ref{fig:janus}, where the 3D objects display multiple heads or duplicated body parts, detracting from the content's quality. Additionally, the process is hindered by the synchronization required between 2D diffusion iteration and NeRF optimization, resulting in a relatively slow optimization process, averaging several hours per scene. While subsequent research, including Magic3D\cite{lin2022magic3d} and Wang et al.~\cite{wang2023prolificdreamer}, has improved aspects like quality and diversity, the Janus issue and the efficiency of the optimization process remain largely under-explored areas.

In this paper, we introduce a novel text-to-3D method capable of generating multi-view consistent 3D content within a reasonable time frame. Key to our approach is the fine-tuning of a 2D diffusion model with explicit orientation control to ensure multi-view aware consistency during Score Distillation Sampling (SDS) optimization. This is achieved by leveraging MVImgNet~\cite{yu2023mvimgnet}, a comprehensive multi-view real object dataset encompassing 6.5 million frames and camera poses. We integrate encoded camera orientation alongside the text prompt, enabling the orientation conditioned 2D diffusion model to provide multi-view constrained 2D priors for supervising NeRF reconstruction.

To accelerate the optimization process, we decouple the back-propagation steps between NeRF parameter updates and the diffusion process. This is implemented using the DDIM solver~\cite{song2020denoising}, which directly resolves the diffusion step to $X_{T-step}$ instead of the conventional $X_{T-1}$. This approach allows for multiple implicit parameter updates per optimization cycle, enhancing efficiency.

Our experiments demonstrate that our method effectively circumvents the multi-head Janus issue while maintaining high quality and variety, outperforming other state-of-the-art methods in terms of time efficiency. As shown in Fig.~\ref{fig:teaser}. In summary, our contributions are threefold:

\begin{figure}[t]
    \centering

    \includegraphics[width=0.9\linewidth]{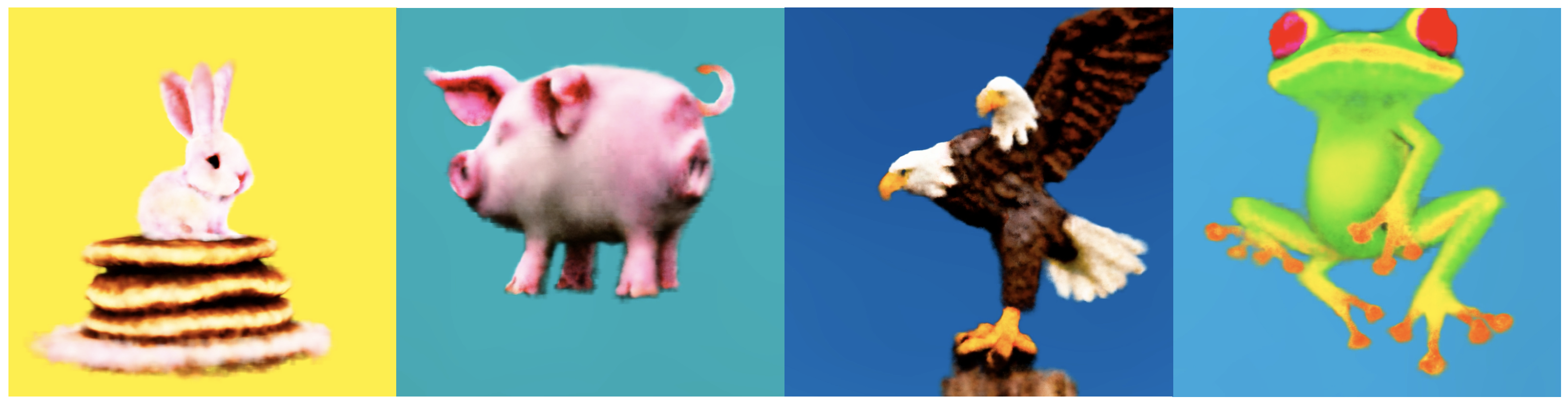}
       \caption{\textbf{Illustration of the Janus Problem}: This figure showcases typical Janus issue manifestations, where the bunny is depicted with three ears, the pig with two noses, the eagle with a pair of heads, and the frog anomalously having three back legs but only one front leg.}
       \label{fig:janus}

\end{figure}

% We propose a camera orientation conditioned 2D diffusion model,
\begin{itemize}
\item In our paper, we introduce a text-to-3D generation method that effectively reduces the multi-head Janus issue and enhances 3D content generation efficiency while maintaining quality and variety.

\item We have developed an advanced 2D diffusion model with explicit orientation control, using MVImgNet data for integrating camera orientation with text prompts. This ensures consistent multi-view accuracy and provides a precise 2D prior for NeRF reconstruction.

\item Our novel optimization strategy significantly speeds up the NeRF reconstruction process. By decoupling NeRF parameter updates from the diffusion process and employing the DDIM solver, we achieve faster and more efficient multiple parameter updates in each optimization cycle.

% Optimized NeRF Optimization Process: Your paper introduces a novel optimization strategy that accelerates the NeRF reconstruction process. By decoupling the back-propagation steps between NeRF parameter updates and the diffusion process and utilizing the DDIM solver for more direct resolution of diffusion steps, your method allows for multiple implicit parameter updates per optimization cycle. This significantly enhances the efficiency and speed of the 3D content generation process.
\end{itemize}

% The score distillation-based text-to-3D methods~\cite{poole2022dreamfusion,lin2022magic3d,wang2022score} assume that maximizing the likelihood of images rendered from arbitrary viewpoints of a NeRF can be translated as maximizing the likelihood of the overall NeRF. Although this is a reasonable assumption, 2D diffusion models lack 3D awareness, which leads to inconsistent and distorted geometry of generated NeRF. To overcome this challenge and ensure NeRF's 3D-consistency, we incorporate 3D awareness into the diffusion model.

% Previous works~\cite{poole2022dreamfusion, wang2022score} attempt this by using the text prompts that roughly describe the camera viewpoint (\textit{e.g.}, ``\textit{side view}''). However, this ad-hoc approach is severely limited: the ambiguity caused by the same text prompt representing a wide range of different pose values leaves NeRF generation vulnerable to geometric inconsistencies. A diffusion model directly conditioned on camera pose value $\pi$ would be an ideal solution.

% Another approach is generate 3D content by lifting 2D images. \YH{} Existing work like zero123 \cite{liu2023zero} generate 3D content from a single image. It utilizes depth estimation method and prior learned from 3D dataset, to infer other views from the given 2D view. This method has inherent limit of single 2D image, that generated model is blurry when on other view angle.

% Therefore, a novel method which produces 3D content that is both high resolution and free from artifacts like Janus problems is much needed.

\section{Related Works}
\label{sec:related_works}

 \subsection{NeRFs for Image-to-3D}
   Neural Radiance Fields (NeRFs), as proposed by Mildenhall et al.~\citep{mildenhall2021nerf}, have revolutionized the representation of 3D scenes through radiance fields parameterized by neural networks. In this framework, 3D coordinates are denoted as \(\mathbf{p} = [x, y, z] \in \mathcal{P}\), and the corresponding radiance values are represented as \(\mathbf{d} = [\sigma, r, g, b] \in \mathcal{D}\).

    NeRFs are trained to replicate the process of rendering frames similar to those produced from multi-view images, complete with camera information. Traditional NeRFs employ a simple mapping of locations \(\mathbf{p}\) to radiances \(\mathbf{d}\) via a function parameterized by a Multilayer Perceptron (MLP). However, recent advancements in NeRF technology have seen the introduction of spatial grids that store parameters, which are then queried based on location. This development, as seen in works by Müller et al.~\citep{muller2022instant}, Takikawa et al.~\citep{takikawa2022variable}, and Chan et al.~\citep{Chan_2022_CVPR}, integrates spatial inductive biases into the model.
    
    We conceptualize this advancement as a \emph{point-encoder} function \(\mathbf{E}_{\theta} : \mathcal{P} \to \mathcal{E}\), with parameters \(\theta\) encoding a location \(\mathbf{p}\) prior to processing by the final MLP \(\mathbf{F} : \mathcal{E} \to \mathcal{D}\). This relationship is mathematically formulated as:
    \begin{equation}\label{eq:point_encoder}
        \mathbf{d} = \mathbf{F}\left(\mathbf{E}_{\theta}\left(\mathbf{p}\right)\right)
    \end{equation}

\subsection{Text-to-Image Generation}\label{sec:text-to-image}
    % Describe our DDM.  Other text-to-image (GAN, whatever) go in related work.
   The widespread availability of captioned image datasets has facilitated the creation of potent text-to-image generative models. Our model aligns with the architecture of recent large-scale methods, as seen in Balaji et al.\citep{balaji2022ediffi}, Rombach et al.\citep{rombach2022high}, and Saharia et al.\citep{saharia2022photorealistic}. We focus on score-matching, a process where input images are modified by adding noise, as outlined by Ho et al.\citep{ho2020denoising} and Song et al.~\citep{song2020score}, and then predicted by the Denoising Diffusion Model (DDM).

    A key feature of these models is their ability to be conditioned on text, enabling the generation of corresponding images through classifier-free guidance, a technique described by Ho et al.~\cite{ho2022classifier}.
    
    In our approach, we employ pre-trained T5-XXL~\citep{JMLR:v21:20-074} and CLIP~\citep{pmlr-v139-radford21a} encoders to produce text embeddings. These embeddings are then used by the DDM, which conditions them via cross-attention with latent image features. Importantly, we repurpose the text token embeddings—denoted as $\textToken$—for modulating our Neural Radiance Field (NeRF), integrating textual information directly into the 3D generation process.

\subsection{Text-to-3D Generation}
    In the realm of Text-to-3D Generation, recent methodologies primarily rely on per-prompt optimization for creating 3D scenes. Methods such as those proposed in Dreamfusion~\cite{poole2022dreamfusion} and Wang et al.~\cite{wang2022score} utilize text-to-image generative models to train Neural Radiance Fields (NeRFs). This process involves rendering a view, adding noise, and then using a Denoising Diffusion Model (DDM), conditioned on a text prompt, to approximate the noise. The difference between the estimated noise and actual noise is used to iteratively update the parameters of the NeRF.

    However, these 2D lifting methods, while impressive in text-to-3D tasks, inherently suffer from 3D inconsistencies, leading to the well-known Janus Problem, as depicted in Fig. \ref{fig:janus}. Various attempts have been made to mitigate this issue. PerpNeg~\cite{armandpour2023re} endeavors to enhance view-dependent conditioning by manipulating the text embedding space. Other strategies focus on stabilizing the Score Distillation Sampling (SDS) process, such as adjusting noise strength~\cite{huang2023dreamtime} and gradient clipping~\cite{hong2023debiasing}. A popular alternative involves leveraging 3D datasets, either by training an additional 2D guidance model on 3D data~\cite{qian2023magic123} or by training a diffusion model to produce multi-view sub-images in a single 2D image~\cite{shi2023mvdream, liu2023syncdreamer, ouyang2023chasing}. However, due to the smaller size and lower texture quality of 3D datasets, outputs from these methods tend to mirror the average quality found in these datasets.~\cite{shi2023mvdream}, outputs from these methods lean towards the average quality in the 3D dataset.

\begin{figure*}
    \centering
       \includegraphics[width=0.9\linewidth]{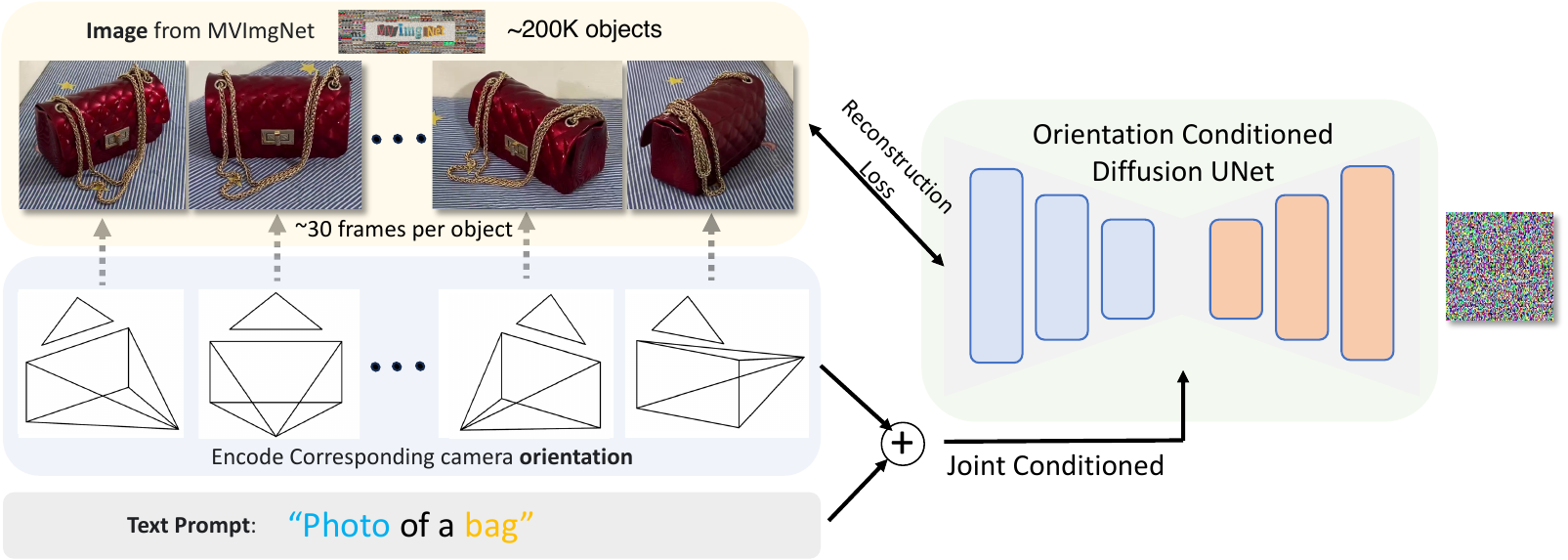}
       \caption{
           \textbf{Overview of camera orientation conditioned Diffusion Model}: This figure illustrates the core components of our innovative model within the \OURS{} pipeline. It highlights how we integrate encoded camera orientations with text inputs, utilizing quaternion forms and sine encoding for precise view angle differentiation. The model, fine-tuned on the real-world MVImgDataset, demonstrates our approach to enhancing 3D consistency and accuracy in NeRF generation, surpassing common limitations found in models trained solely on synthetic 3D datasets.
           }
       \label{fig:arch_2d}
       
\end{figure*}

\section{Method}

% \subsection{Motivation and Overview}

% We now present xxxxxxxxx. By using the 3D ddddd. 3.1 posconditioned.....  3.2 step for generation 3.3 decoupled
Our proposed method is structured into three critical sections, each essential to our innovative approach in text-to-3D generation: First, Section~\ref{sec:Orientation_difussion} delves into the camera orientation conditioned diffusion model. Here, we introduce a diffusion model that adeptly integrates camera orientation, significantly enhancing the precision and consistency of our 3D content generation from various viewpoints. Second, Section~\ref{sec:Text_to_3D_generation} covers the Text-to-3D Generation process. In this part, we detail how our method effectively translates text descriptions into detailed 3D models. We focus on overcoming specific challenges, such as the multi-head Janus issue, and on enhancing the overall efficiency of the generation process. Lastly, Section~\ref{sec:decoupled_back_prop} presents our Decoupled Back Propagation approach. This section is dedicated to outlining our novel strategy for optimizing the NeRF model. We emphasize our method's capacity to improve both the speed and the diversity of 3D generation by employing an advanced back-propagation technique.

% In the following, we describe the camera pose encoding, pose-conditioned diffusion model, and how it could be applied on text-to-3d generation.
%\begin{figure}[!h]
%    \centering
%    \includegraphics[width=\linewidth]{img/camera_encoding.pdf}
%       \caption{\textbf{Camera Encoding}. }
%       \label{fig:camera_encoding}
%\end{figure}

\subsection{Camera Orientation Conditioned Diffusion Model}
\label{sec:Orientation_difussion}

Score distillation-based text-to-3D methods, such as those proposed in Poole et al.\cite{poole2022dreamfusion}, Lin et al.\cite{lin2022magic3d}, and Wang et al.~\cite{wang2022score}, operate under the assumption that maximizing the likelihood of images rendered from arbitrary viewpoints of a Neural Radiance Field (NeRF) is tantamount to maximizing the NeRF's overall likelihood. While this assumption is rational, the 2D diffusion models these methods employ lack 3D awareness, often resulting in inconsistent and distorted geometry in the generated NeRF. To address this limitation and ensure the reconstructed NeRF maintains 3D consistency, our approach integrates 3D awareness into the 2D image diffusion generation model.

Prior works, including those by Poole et al.\cite{poole2022dreamfusion} and Wang et al.\cite{wang2022score}, have attempted to incorporate a sense of 3D awareness by using text prompts that vaguely describe the camera viewpoint (for example, 'side view'). However, this method is inherently flawed due to its ad-hoc nature. The ambiguity inherent in using text prompts that can represent a wide range of different pose values makes NeRF generation susceptible to geometric inconsistencies. To overcome this, we propose an approach that explicitly controls our 2D diffusion image generation model by inputting encoded camera orientations.
% which is directly conditioned on camera pose value $p$ would be beneficial.

In our proposed method, we utilize a standard Latent Diffusion model~\cite{metzer2023latent} with camera orientation control, which can be mathematically represented as follows:

\begin{equation}
p(\mathbf{x} | \theta) = \prod_{t=1}^{T} p(\mathbf{x} | \theta, t, c'),
\end{equation}

where $p(\mathbf{x} | \theta)$ denotes the data distribution parametrized by $\theta$, $\mathbf{x}$ represents the data (e.g., an image), and $t$ signifies the number of diffusion steps. The conditional distribution of the data at step $t$ is given by $p(\mathbf{x} | \theta, t, c')$, where $c'$ is a concatenated feature combining the CLIP feature $c$ from the text input and the encoded orientation $SE(q)$, both having the same shape. Our method involves adding $p$ to $c$ as a residual, resulting in $c' = c + p$.

To encode camera orientation, we first convert the camera orientation to a quaternion form $q$ due to the numerical discontinuity of the rotation matrix. To facilitate smooth interpolation between camera poses, we apply a sine encoding $\left\{{{SE}}_i(q), i=1,...,7\right\}$, where $i$ represents the dimension. The sine encoding process is as follows:

\begin{equation}
\text{{SE}}_i(q) = \sin\left(q \cdot i \right)
\end{equation}

This encoding allows an intermediate quaternion between two poses $q_1$ and $q_2$ to be represented as a simple linear interpolation, which is straightforward for the network to learn and generalize:

\begin{equation}
\text{{SLerp}}(q_1, q_2, t) = \frac{{\sin((1 - t)\omega)}}{{\sin(\omega)}} q_1 + \frac{{\sin(t\omega)}}{{\sin(\omega)}} q_2
\end{equation}

where $\omega$ is the angle between the two quaternions, determined as $\omega = \cos^{-1}(q_1 \cdot q_2)$.

We evaluated our camera encoding scheme against alternative approaches like relative position encoding~\citep{singer2022make}, rotary embeddings~\citep{su2021roformer}, and the plain rotation matrix, concluding that the quaternion form with sine encoding most effectively distinguishes between view angles.

With the above setup, we fine-tune the diffusion model as shown Fig.~\ref{fig:arch_2d}, and using the following loss function:

\begin{align}
\min_{\theta}; \mathbb{E}{z \sim \mathcal{E}(x), t, \epsilon \sim \mathcal{N}(0, 1)}||\epsilon - \epsilon{\theta}(x_t, t, c, SE(q))||_2^2.
\end{align}

Here, $x_t$ is the noisy image generated from random noise $\epsilon$, $t$ is the timestep, $c$ is the condition, $\mathbf{p}$ is the camera condition, and $\epsilon_\theta$ is the view-conditioned diffusion model.

After training the model $\epsilon_\theta$, the inference model $f$ can generate an image through iterative denoising from a Gaussian noise image, as described in Rombach et al.~\cite{rombach2022high}. This process is conditioned on the parameters $(t, c, p)$.

Fine-tuning pre-trained diffusion models in this manner equips them with a generalized mechanism for controlling camera viewpoints, allowing for extrapolation beyond the objects encountered in the fine-tuning dataset. Unlike methods that rely on 3D datasets, such as those discussed in Liu et al.\cite{liu2023zero} and Shi et al.\cite{shi2023mvdream}, our model benefits from fine-tuning on the real-world captured MVImgDataset~\cite{yu2023mvimgnet}. This approach avoids the common issues of reduced content richness and texture quality associated with models trained exclusively on 3D datasets.

Camera poses in MVImgNet~\cite{yu2023mvimgnet} are estimated using COLMAP~\cite{schoenberger2016sfm}. Despite being captured by different annotators, the normalization term used in COLMAP ensures that the camera poses are canonical across different captures.

\begin{figure*}[!h]
    \centering
       \includegraphics[width=0.85\linewidth]{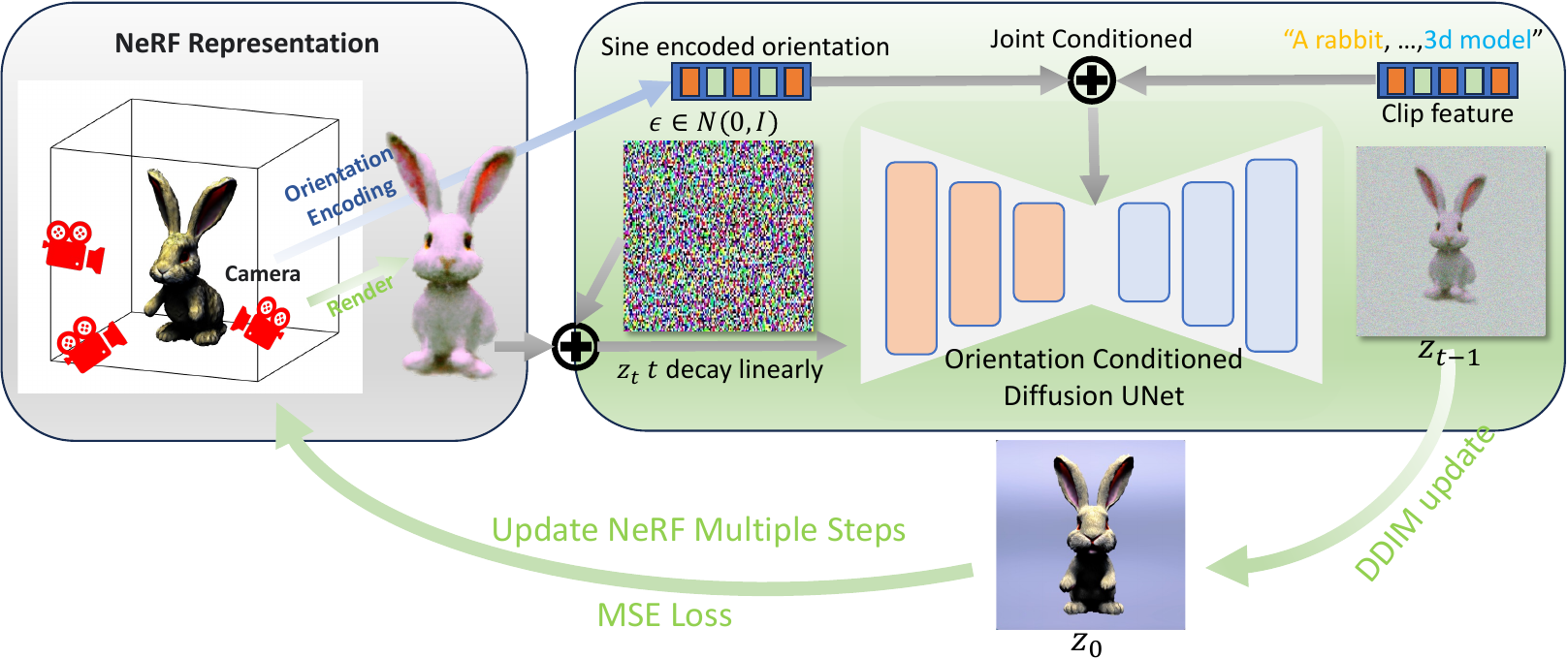}
       \caption{
           \textbf{Overview of Text-to-3D Generation Methodologies}: This figure succinctly illustrates our dual approach in applying multi-view diffusion models for 3D generation. It highlights the use of multi-viewpoint images for sparse 3D reconstruction, alongside our focus on employing an orientation-conditioned diffusion model for Score Distillation Sampling (SDS). We figure showcases our innovative SDS adaptation, where traditional models are replaced with our orientation conditioned diffusion model, seamlessly integrating camera parameters and text prompts for enhanced 3D content generation.
           }
       \label{fig:arch}
\end{figure*}

\subsection{Text to 3D generation}
\label{sec:Text_to_3D_generation}
We explore two primary methodologies for applying a multi-view diffusion model to 3D generation. The first method involves using images generated from multiple viewpoints as inputs for sparse 3D reconstruction. This technique, while effective, requires high-quality images and poses, which has inspired our work described in Section \ref{sec:decoupled_back_prop}.

The second method leverages the multi-view diffusion model as a prior for Score Distillation Sampling (SDS). This approach is more robust to imperfections in images and tends to produce superior output quality. In our work, we focus on this latter method as described in Fig.\ref{fig:arch}, modifying existing SDS pipelines by replacing the Stable Diffusion model with our pose-conditioned diffusion model. Instead of utilizing direction-annotated prompts as in Dreamfusion~\citep{poole2022dreamfusion}, we employ original prompts for text embedding extraction and incorporate camera parameters as inputs.

Furthermore, we use our orientation-conditioned diffusion model to initialize the variational model in Variational Score Distillation (VSD), since both models use camera pose as a conditioning element. This integration allows for a more nuanced and accurate 3D generation process, capitalizing on the strengths of our developed diffusion model.

% We use DmTet as the second stage refinement.

\subsection{Decoupled Back Propagation}
\label{sec:decoupled_back_prop}

\begin{figure}[t]
    \centering

    %\includegraphics[width=\linewidth]{img/sampler.pdf}
    %   \caption{\textbf{SDS Sampling}. }

    \includegraphics[width=\linewidth]{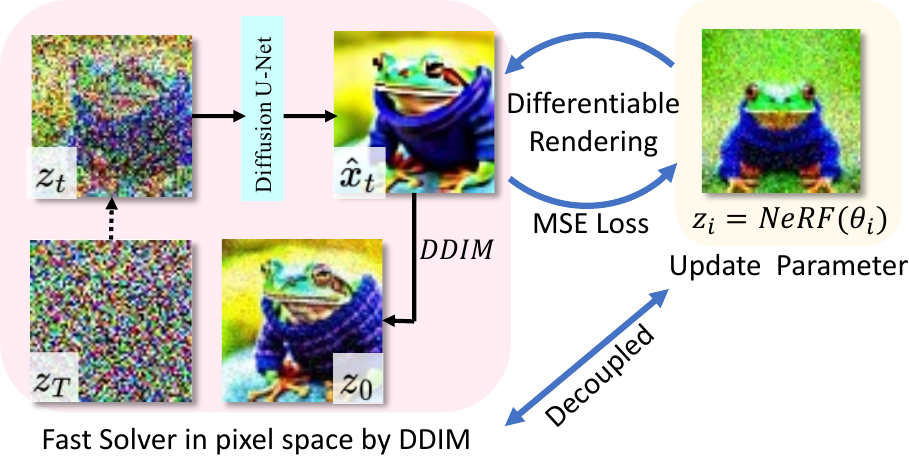}

       \caption{\textbf{Decoupled Sampling}: We summarizes our approach to enhance NeRF optimization speed, highlighting the shift from uniform sampling to targeted reduction in $T$ steps and the use of the DDIM solver for efficient computation, thereby improving both the diversity and texture quality in 3D model generation. }
       
       \label{fig:sampler}

\end{figure}

The current methodologies still grapple with slow optimization speeds. As highlighted in Prolificdreamer~\cite{wang2023prolificdreamer}, the mode-seeking nature of Score Distillation Sampling (SDS) can lead to over-saturation and reduced diversity. Prolificdreamer suggests incorporating variational constraints to address this, but this addition further decelerates the learning process.

Recent advancements in Ordinary Differential Equation (ODE) samplers for diffusion models, like DDIM~\cite{song2020denoising} and DPM-Solver~\cite{lu2022dpmsolver}, offer rapid evaluation of 2D diffusion models by potentially solving the ODE in a single step.

In our approach to decoupled back-propagation, we implement a series of strategic steps to enhance the efficiency of the process. Initially, we move away from uniform sampling, opting instead to strategically decrease the choice of $T$. Following this, as shown in Fig.~\ref{fig:sampler}, we sample a random camera pose and render a corresponding 2D view of the NeRF. Then, using the DDIM solver, we directly compute $X_{T-step}$, diverging from the traditional method of calculating $X_{T-1}$. This step size is not static; it's a hyperparameter that we methodically reduce during the course of training. Finally, once $X_{T-step}$ is determined, we proceed to optimize the NeRF at this specific camera location, employing a Mean Squared Error (MSE) loss for this purpose.

The DDIM solver formula is expressed as:
\begin{equation}
x_t = x_0 + \sqrt{2 \beta} \sum_{i=1}^{t} \sigma_i \epsilon_i
\end{equation}

where \( x_t \) denotes the image at time \( t \), \( x_0 \) signifies the initial image, \( \beta \) is the parameter governing the diffusion process, \( \sigma_i \) represents the standard deviation of the noise at step \( i \), and \( \epsilon_i \) is the noise sample at the same step.

This method circumvents the problematic SDS loss and instead employs standard MSE loss to train NeRF, which results in improved diversity and texture quality. Towards the end of training, as the step size becomes smaller, the quality aligns closely with that of a 2D diffusion solver.

A significant advantage of this method is the considerable acceleration of the training process. The stable diffusion 2.1 model having 890 million parameters, are substantial in size compared to the much smaller NeRF models, which have only about 2.3 million parameters.

Decoupled back propagation requires fewer evaluations of the diffusion model, thus reducing the frequency of optimizing NeRF from different viewing angles. This efficiency is achievable only when the 2D diffusion model can precisely generate views from specified locations, making the most of its advanced capabilities.

\begin{figure*}[h!]
\hspace{-5em}
\centering
\begin{subfigure}[b]{0.9\textwidth}
    \centering
\begin{tabular}{lll}
\toprule
\raisebox{-0.15\height}{\rotatebox{90}{SD + Text Suffix}}
& \includegraphics[width=0.5\linewidth]{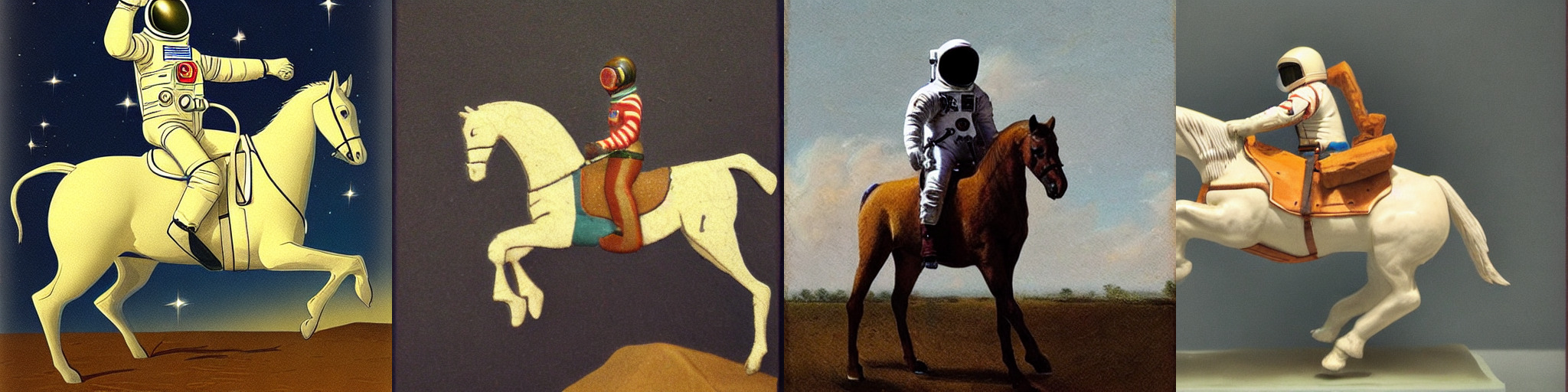} &
\includegraphics[width=0.5\linewidth]{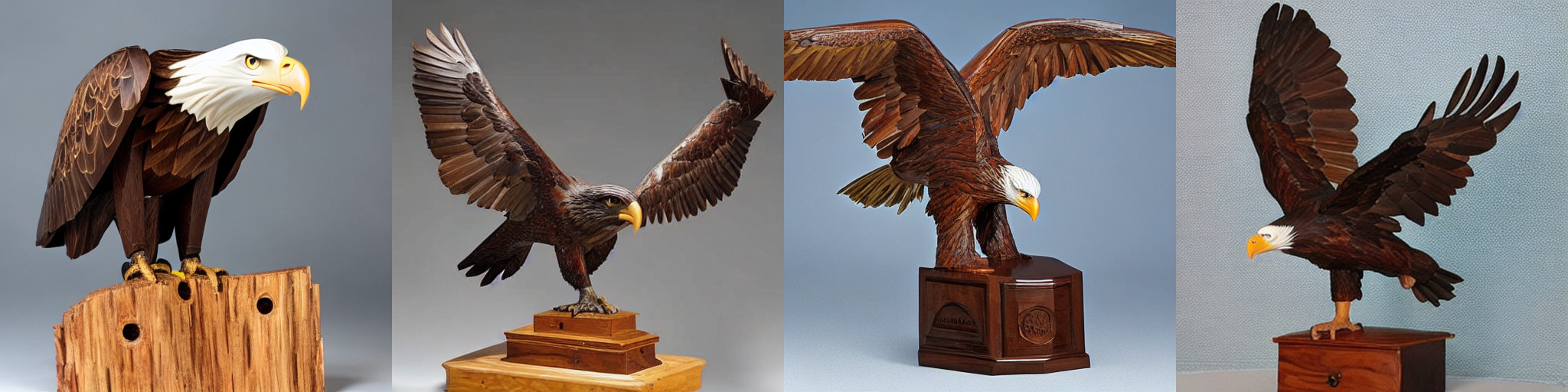}\\
\raisebox{0.1\height}{\rotatebox{90}{MVDream}}
& \includegraphics[width=0.5\linewidth]{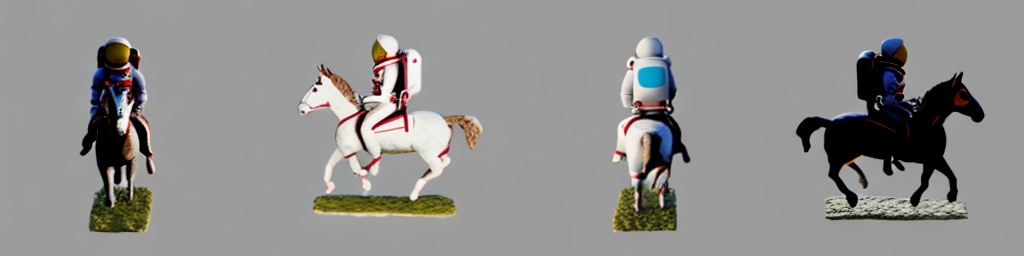} &
\includegraphics[width=0.5\linewidth]{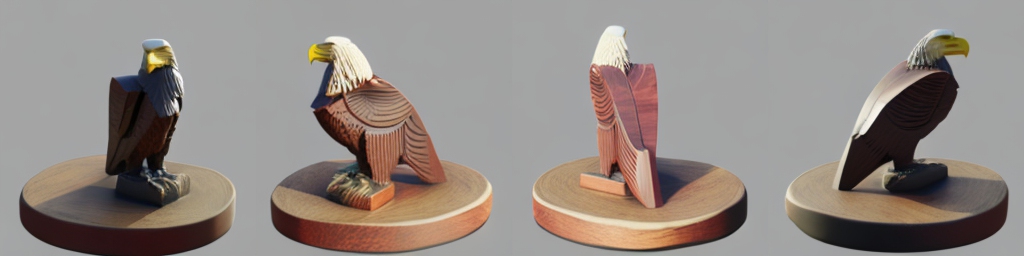} \\
\raisebox{1.2\height}{\rotatebox{90}{Our}}
& \includegraphics[width=0.5\linewidth]{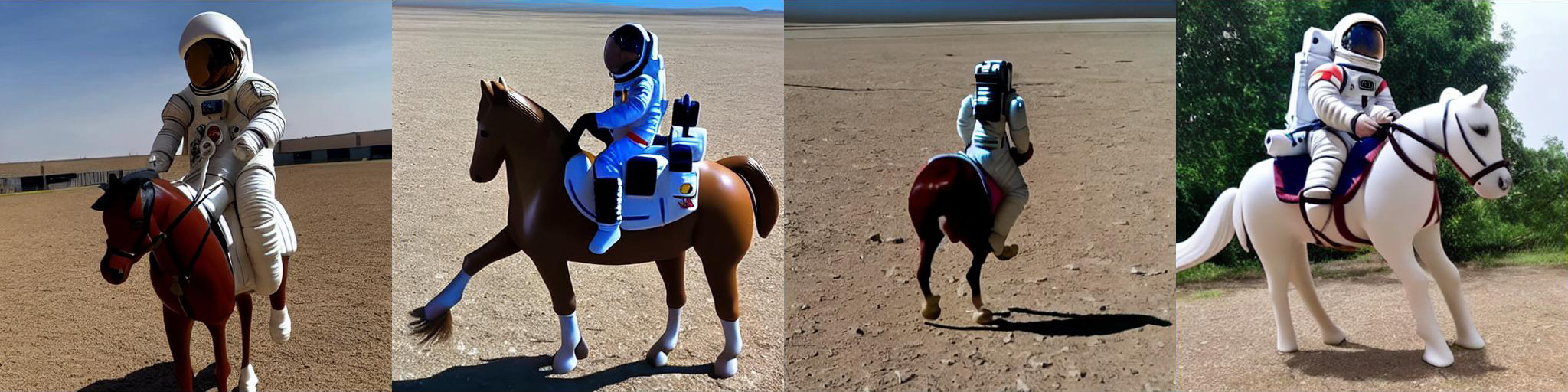} &
\includegraphics[width=0.5\linewidth]{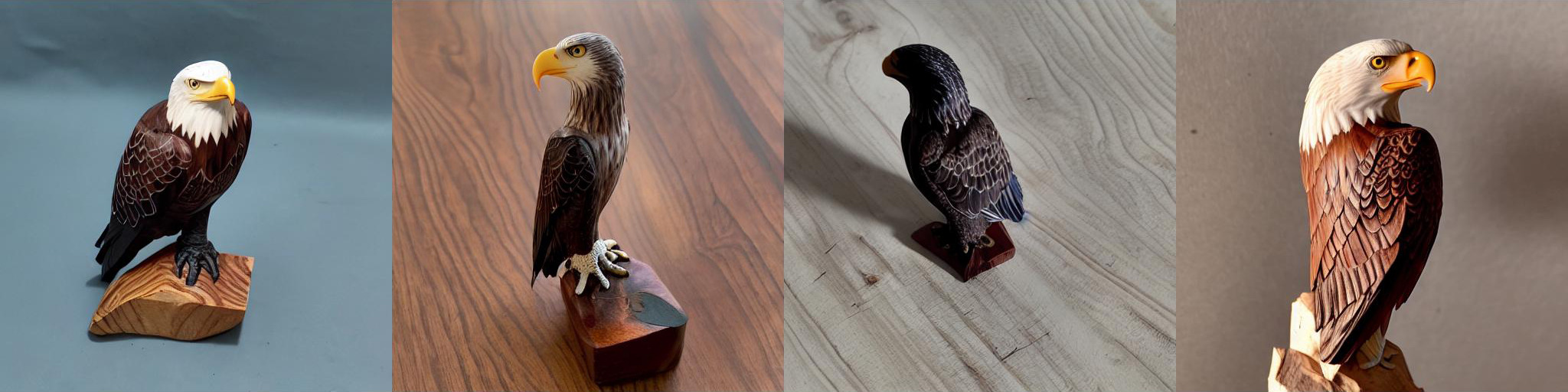} \\
\midrule
& {\centering An astronaut riding a horse}
& {\centering A bald eagle carved out of wood }\\
\midrule

\raisebox{-0.15\height}{\rotatebox{90}{SD + Text Suffix}}
& \includegraphics[width=0.5\linewidth]{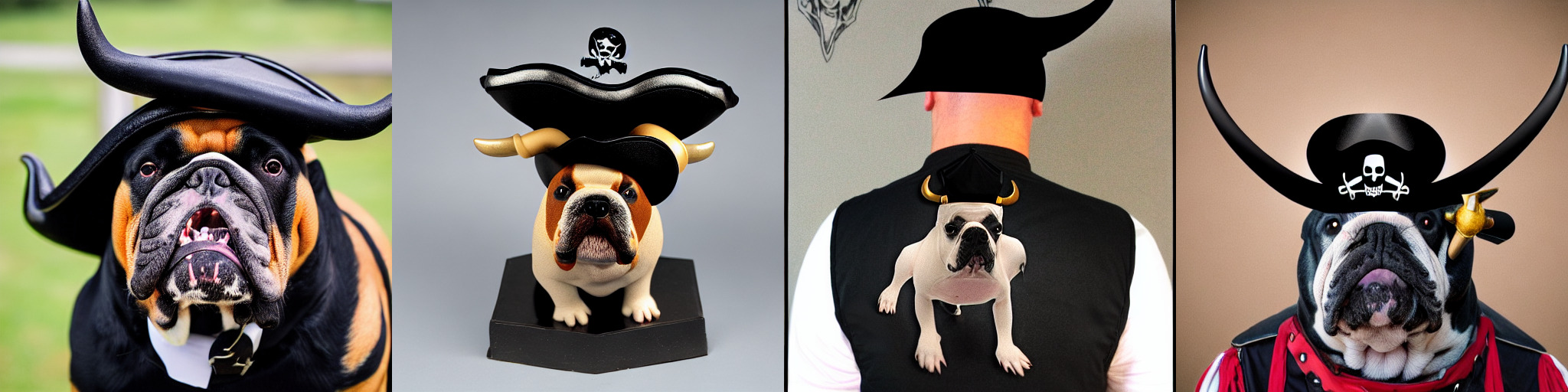} &
\includegraphics[width=0.5\linewidth]{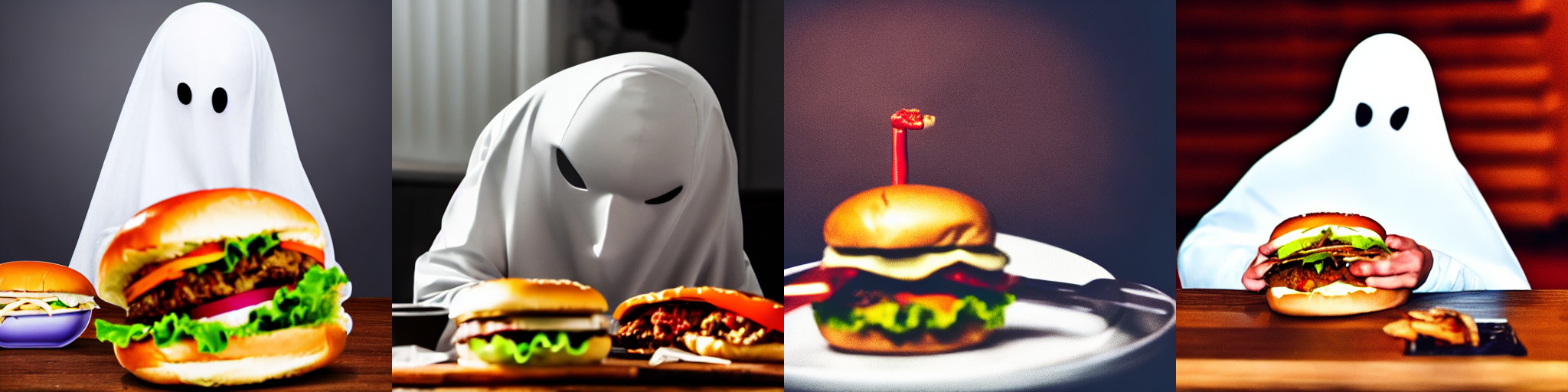}\\
\raisebox{0.1\height}{\rotatebox{90}{MVDream}}
& \includegraphics[width=0.5\linewidth]{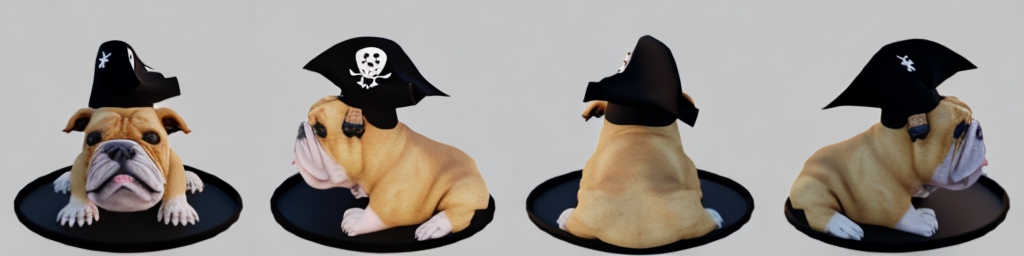} &
\includegraphics[width=0.5\linewidth]{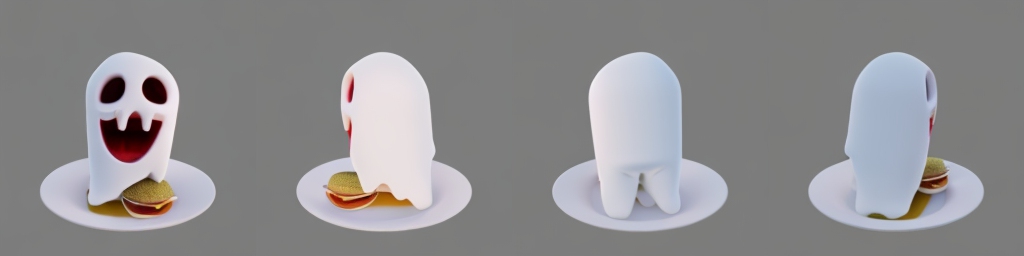} \\
\raisebox{1.2\height}{\rotatebox{90}{Our}}
& \includegraphics[width=0.5\linewidth]{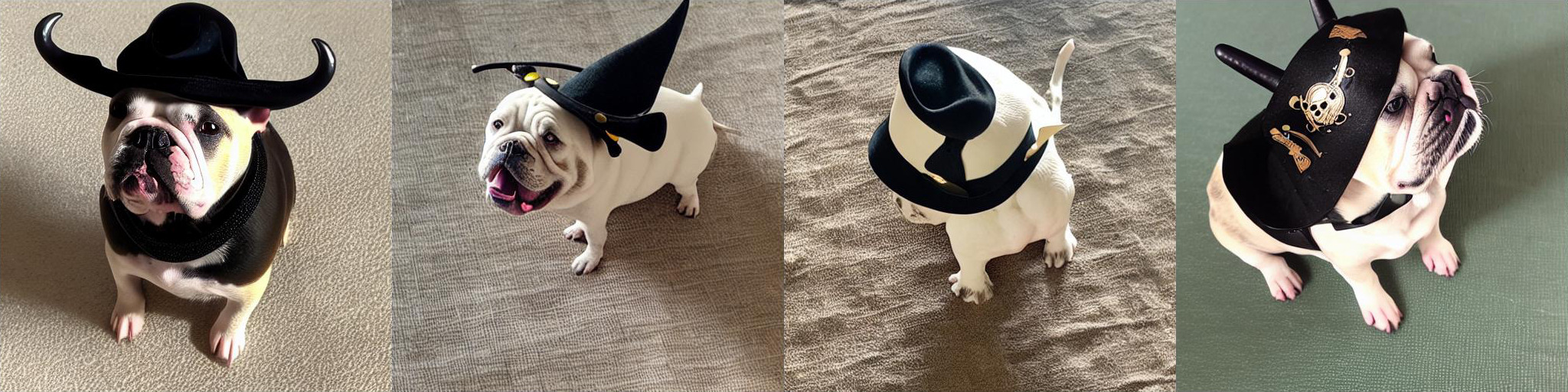} &
\includegraphics[width=0.5\linewidth]{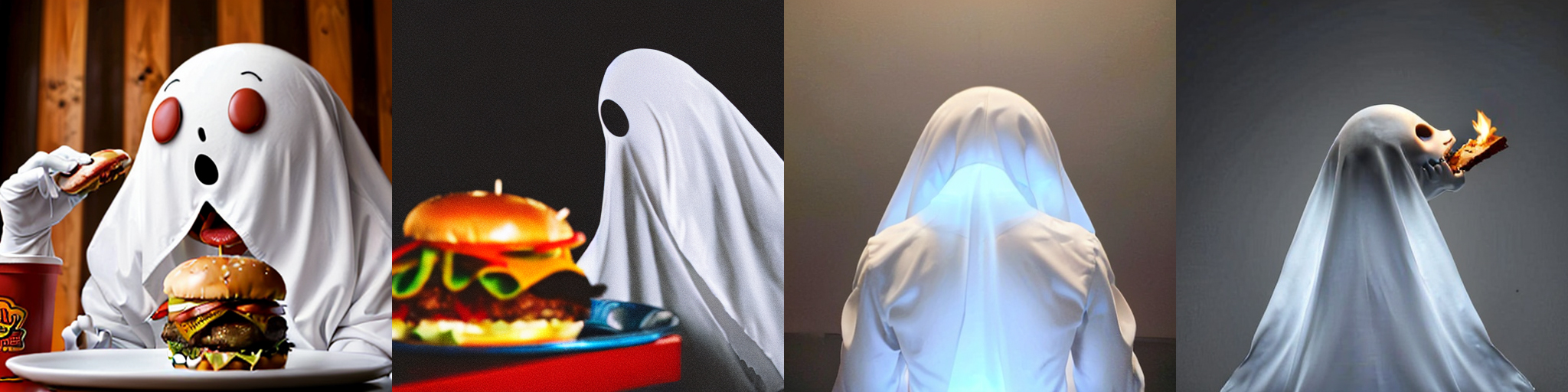} \\
\midrule
& A bull dog wearing a black pirate hat
& A DSLR photo of a ghost eating a hamburger
\\\bottomrule

\end{tabular}
\end{subfigure}
    \caption{\textbf{Qualitative Results of Orientation-Controlled Image Generation}: We showcases the comparative results of our method against the text suffix-based approach and MVDream~\cite{shi2023mvdream}. While the text-based method yields images with inconsistent orientations, both MVDream and our model exhibit precise control over viewing angles. Notably, our method outperforms MVDream by generating photorealistic images with richer textures and greater variety, highlighting its effectiveness in producing high-quality, orientation-specific imagery.}
    \label{fig:2d_results}
    \vspace{-2em}
\end{figure*}

\section{Experiments}
In this section, we begin by reviewing the training details (Section~\ref{sec:training}). We then assess our orientation conditioned diffusion model in the context of 2D image generation (Section~\ref{sec:2d}), followed by an evaluation of its performance in 3D object generation using a 2D to 3D lifting method (Section~\ref{sec:3d}). Subsequently, we provide a quantitative evaluation of the generation quality and speed (Section~\ref{sec:quantitative}) . Due to space limit, ablation study of Decoupled  Back Propagation is left in Appendix.

\subsection{Training Details}
\label{sec:training}
Our model, an orientation-controlled diffusion variant, builds upon the Stable Diffusion 2.1 framework. It is fine-tuned using the MVImgNet dataset~\cite{yu2023mvimgnet}, which comprises 6.5 million frames across 219,188 videos featuring objects from 238 classes. The fine-tuning process is carried out over 2 epochs. For the 2D to 3D lifting aspect, we adopt methodologies from Dreamfusion~\cite{poole2022dreamfusion, stable-dreamfusion}. We have implemented our approach within the Threestudio~\cite{threestudio} framework, which is also employed for replicating baseline models.

\subsection{Orientation-Controlled Image Generation}
\label{sec:2d}
Our orientation-controlled diffusion model is specifically trained on an orientation-annotated dataset. To verify its effectiveness, we initially focus on generating open-domain objects at specified camera orientations.

We utilized the same prompts as those in MVDream~\cite{shi2023mvdream}. For each prompt, four images are generated, each with the same elevation but differing in azimuth by 90 degrees. Both MVDream~\cite{shi2023mvdream} and our method employ orientation quaternions for this purpose.

As a comparative measure, we also incorporate a text-based view conditioning method. This approach involves adding text suffixes like \textit{front view, side view, back view} to the text prompt, corresponding to the camera orientation. This method is a common feature in text-to-3D pipelines, as seen in works like Threestudio and Stable-Dreamfusion~\cite{threestudio, stable-dreamfusion}.

In Figure~\ref{fig:2d_results}, we display the results of these generative approaches. The text suffix-based method tends to produce images in various orientations, rather than consistent ones, underscoring its limitations. Conversely, both MVDream and our model demonstrate reliable control over viewing orientation. However, while MVDream tends to produce 2D images with simplified textures and limited variety, our method achieves photorealistic imagery. This distinction is crucial as it directly impacts the quality of text-to-3D generation.

MVDream generate 4-views simultaneously, while our method generate 4-views independently, it's not an apple-to-apple comparison to evaluate 4-view consistency. Considering text-to-3d training require around 10,000 iterations, consistency within a 4-view pair is not critical. On the other hand, our method has clear advantage on texture quality and content richness.

\begin{figure*}[h]
\centering
\begin{subfigure}[b]{0.9\textwidth}
    \centering
    \setlength\tabcolsep{0px}
    \renewcommand{\arraystretch}{0.0}
    \newcommand{\degr}[1]{\small#1$^{\circ}$}
    \newcolumntype{Y}{>{\centering\arraybackslash}X}
    \captionsetup[subfigure]{labelformat=empty}
    
    \begin{tabularx}{1.0\linewidth}{p{0.5cm} YYYY p{2px} YYYY}
    & \degr{0} & \degr{45} &  \degr{90} & \degr{135} &  
    & \degr{0} & \degr{45} &  \degr{90} & \degr{135} \\
    \end{tabularx}
    \setlength\tabcolsep{1px}
    \begin{tabularx}{1.0\linewidth}{c cc}
    {\rotatebox{90}{\small Dreamfusion}} &
    \includegraphics[width=0.5\linewidth]{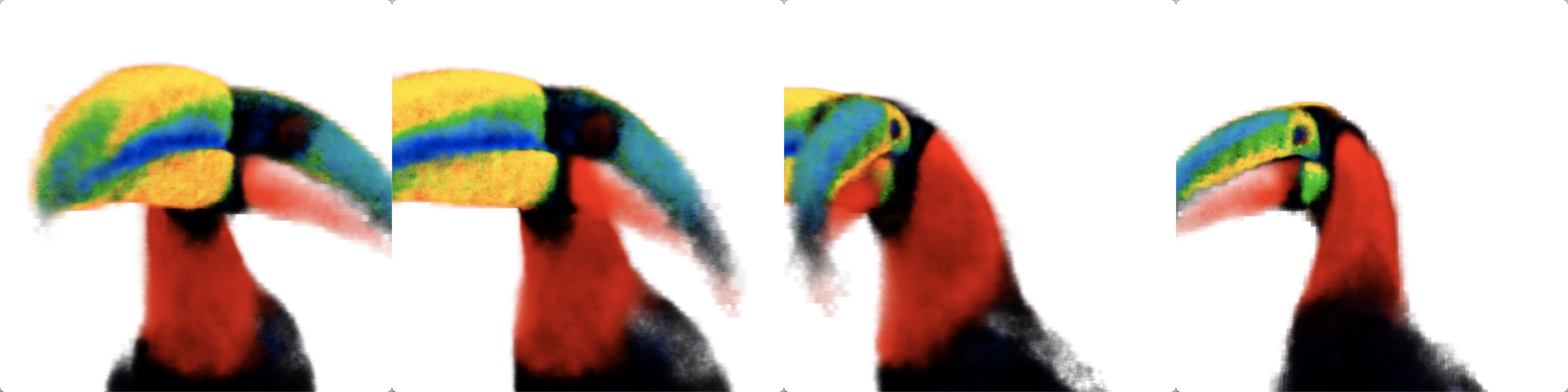} &
    \includegraphics[width=0.5\linewidth]{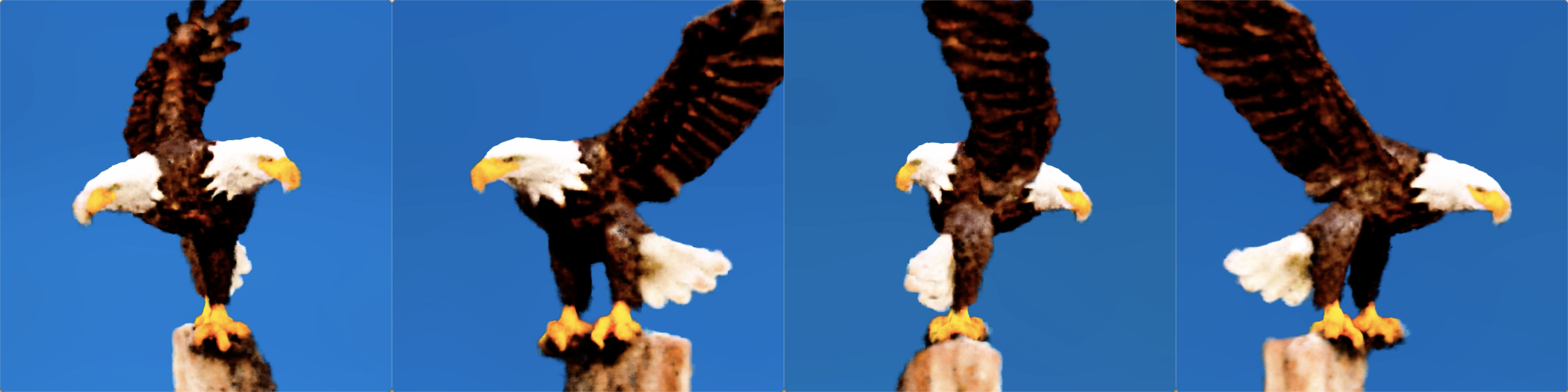}\\
    {\rotatebox{90}{\small ProlificDreamer}} &
    \includegraphics[width=0.5\linewidth]{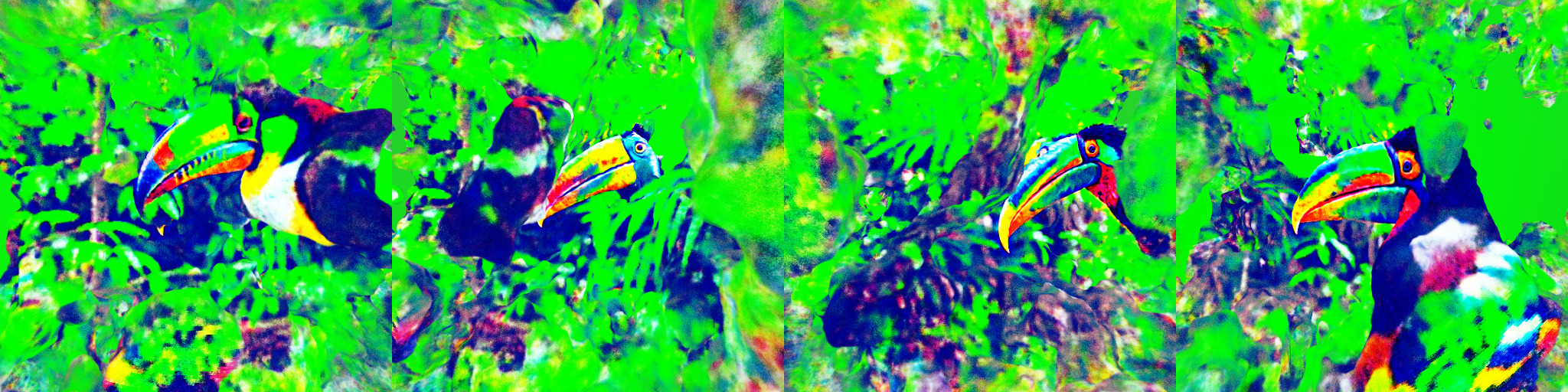}&
    \includegraphics[width=0.5\linewidth]{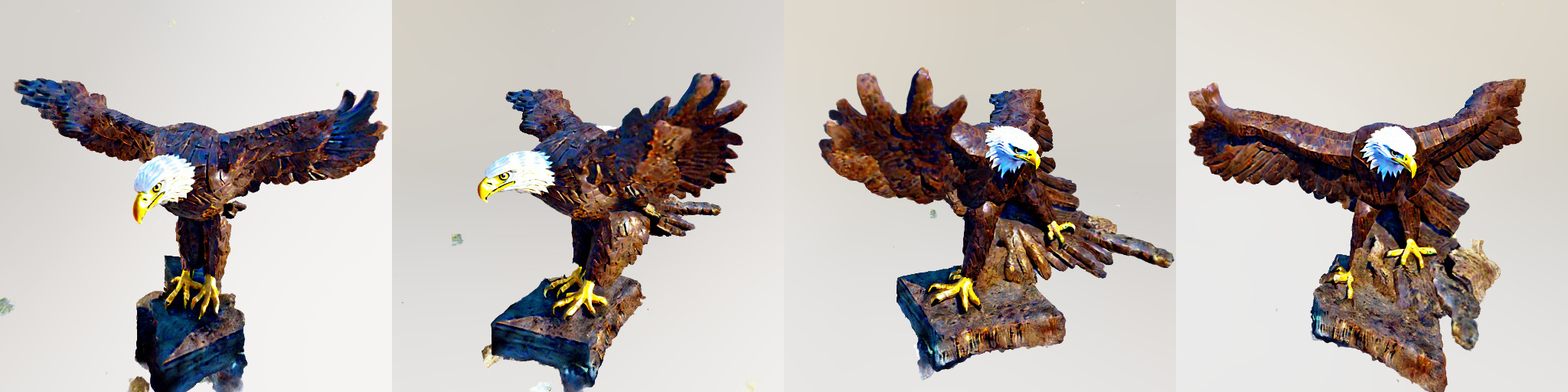}\\
    \raisebox{0.2\height}{\rotatebox{90}{MVDream}}&
    \includegraphics[width=0.5\linewidth]{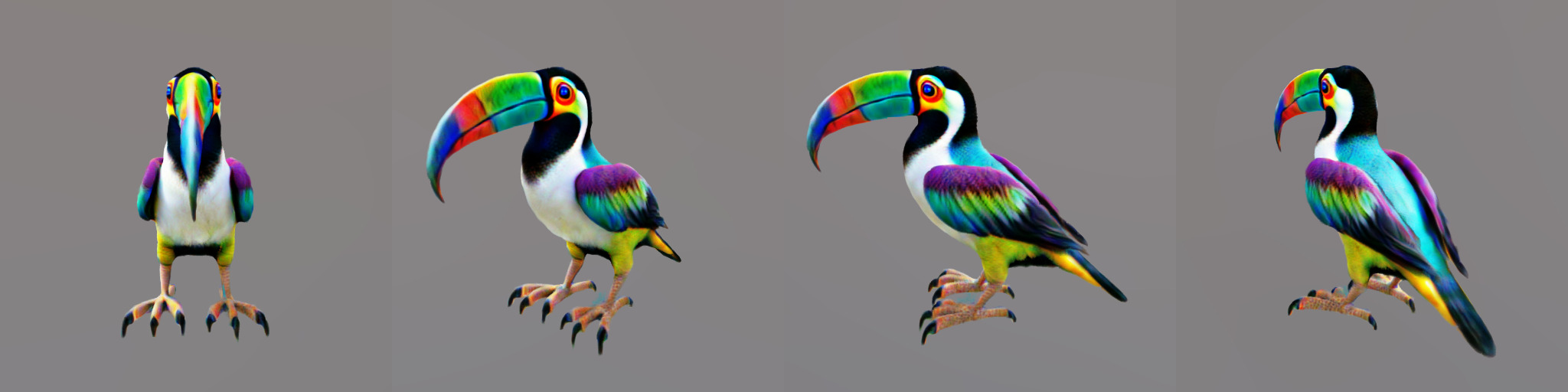} &
    \includegraphics[width=0.5\linewidth]{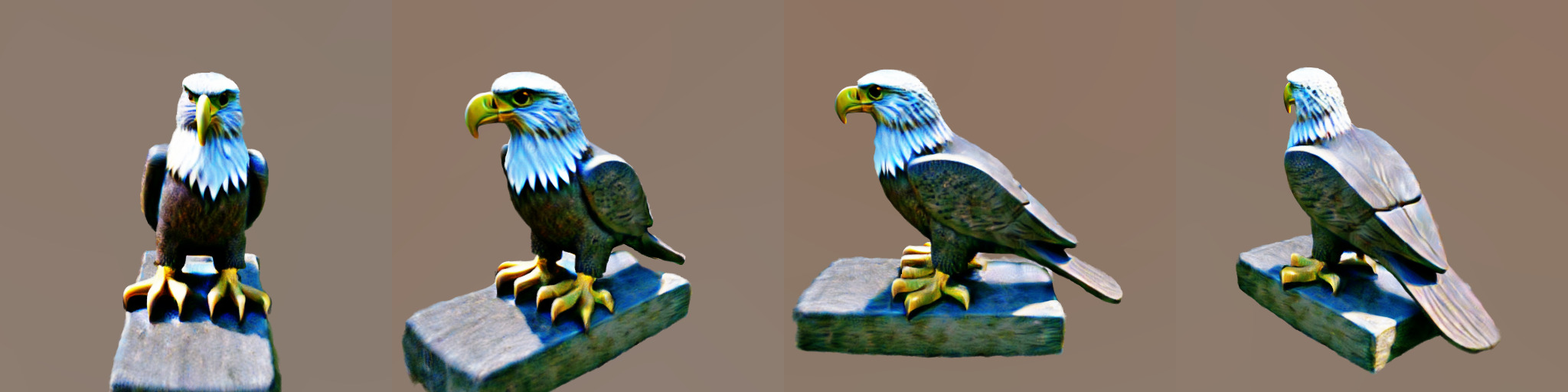}\\
    %{\rotatebox{90}{\tiny Our-SDS}} &
    %\includegraphics[width=0.5\linewidth]{img/3d/our_sds/toucan.jpg} &
    %\includegraphics[width=0.5\linewidth]{img/3d/our_sds/eagle.jpg} \\
    %{\rotatebox{90}{\tiny Our-SDS-Normal}} &
    %\includegraphics[width=0.5\linewidth]{img/3d/our_sds_normal/toucan.jpg} &
    %\includegraphics[width=0.5\linewidth]{img/3d/our_sds_normal/eagle.jpg}  \\
     
    \raisebox{1.5\height}{\rotatebox{90}{\small Our}} &
    \includegraphics[width=0.5\linewidth]{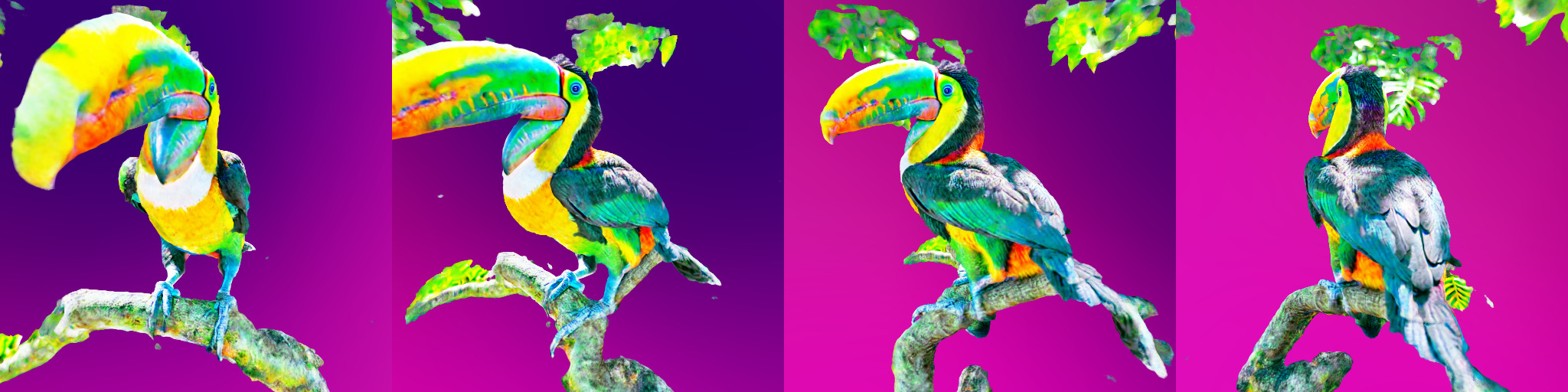} &
    \includegraphics[width=0.5\linewidth]{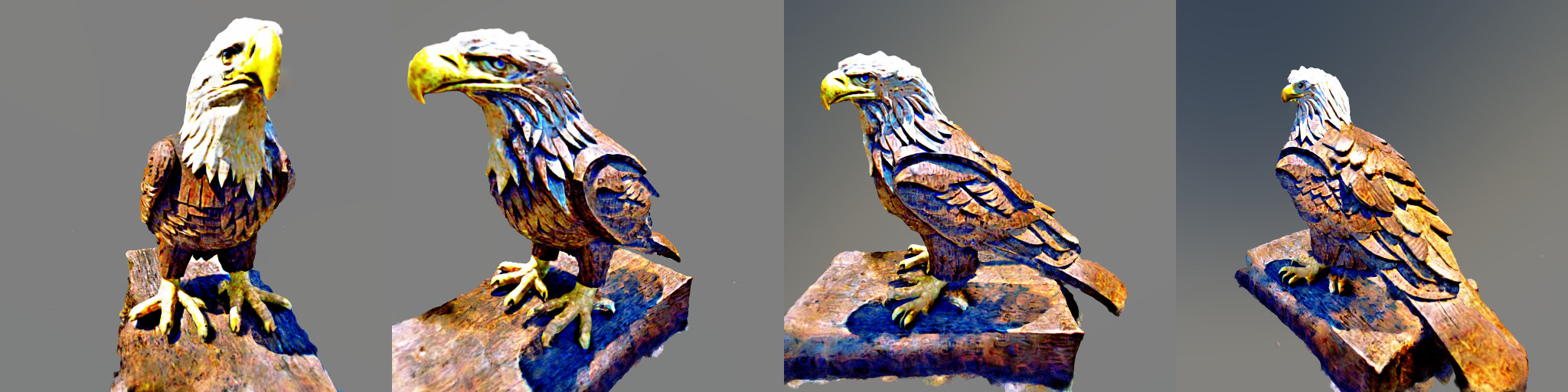}\\
    \raisebox{0.2\height}{\rotatebox{90}{\small Our-Normal}} &\includegraphics[width=0.5\linewidth]{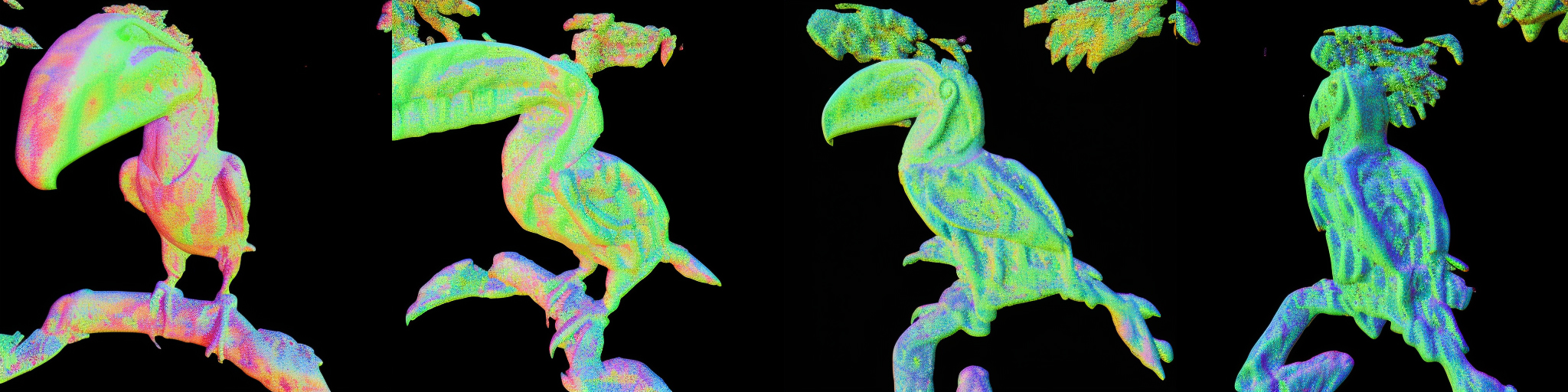}
    &\includegraphics[width=0.5\linewidth]{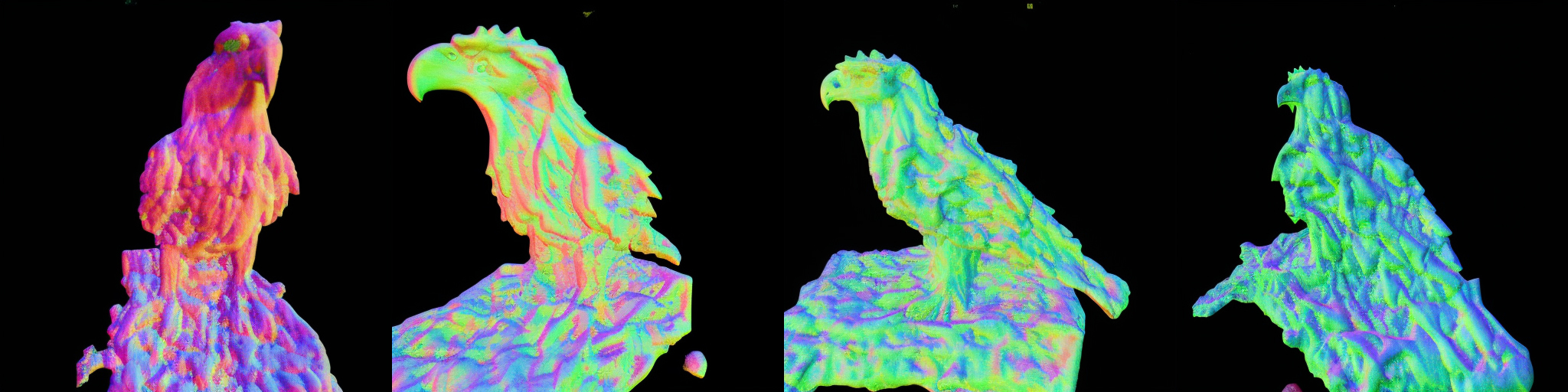}\\
    \midrule\\
    & A colorful toucan with a large beak
    & a bald eagle carved out of wood\\
    \bottomrule
    \end{tabularx}
\end{subfigure}
        \caption{\textbf{Qualitative experiment of text to 3D generation results.} Each row represents a method and each column represents a azimuth angle used to render the object. }
            \label{fig:3d_results}
\end{figure*}

\subsection{Orientation-Controlled Diffusion for Text to 3D}
\label{sec:3d}

Following the approach in Dreamfusion~\cite{poole2022dreamfusion,stable-dreamfusion}, we employ the NeRF architecture with Instant-NGP representation~\cite{muller2022instant} for lifting 2D features to 3D. Our comparison includes prominent baselines like Dreamfusion~\cite{poole2022dreamfusion}, Prolificdreamer~\cite{wang2023prolificdreamer}, and MVDream~\cite{shi2023mvdream}. In Figure.~\ref{fig:3d_results}, we display our text-to-3D generation results. Dreamfusion~\cite{poole2022dreamfusion} tends to suffer from the Janus problem, as seen in the duplication of heads in both the toucan and eagle models. Prolificdreamer~\cite{wang2023prolificdreamer} achieves impressive texture quality in individual views, but viewed as a coherent 3D object, it exhibits an even more pronounced Janus issue. MVDream~\cite{shi2023mvdream} maintains good 3D consistency, yet it falls behind in terms of content richness and texture quality compared to other methods. Our method stands out by being free from the Janus problem and providing exceptional detail and texture quality.

\subsection{Quantitative Evaluation}
\label{sec:quantitative}
Evaluating the quality of text-to-3D generation is challenging due to the lack of ground truth 3D scenes corresponding to text prompts. CLIP R-Precision~\cite{poole2022dreamfusion} is one metric, assessing retrieval accuracy through joint embedding of 2D images and text. However, high image-text matching scores don't necessarily imply image consistency.

A-LPIPS~\cite{hong2023debiasing}, calculating average LPIPS~\cite{lpips} between adjacent images, is another metric. But frame-wise similarity doesn't guarantee view consistency, and the Janus problem, characterized by repetitive patterns, could misleadingly improve this metric. An object appearing identical from every angle would score highly on A-LPIPS, despite lacking view diversity.

\begin{table}[b]

\footnotesize  
\centering
\scalebox{0.9}{
\begin{tabular}{l c c c} 
\toprule
 Method & R-Precision $\uparrow$ & $\text{A-LPIPS}_\text{Alex} \downarrow$  & $\text{Zero123}_{grad} \downarrow$ \\
\midrule
Dreamfusion\cite{poole2022dreamfusion} & 58.8\% & \textbf{0.0250} & 0.0873 \\
Prolificdreamer\cite{wang2023prolificdreamer} & \textbf{63.4\%} & 0.0688 & 0.1412\\
MVDream\cite{shi2023mvdream} & 48.4\% & 0.0311 & \textbf{0.0092} \\
Ours & 55.5\% & 0.0415 & 0.0107 \\
\bottomrule
\end{tabular}
}
\caption{\textbf{Comparision of R-Precision, A-LPIPS, and $\text{Zero123}_{\text{grad}}$}. Results are calculated on 64 prompts dataset. }
\label{tab:zero123_grad}
\end{table}

To address these limitations, we developed a 3D topology-aware metric, leveraging models like Zero123\cite{liu2023zero}, MVDream\cite{shi2023mvdream}, and our \OURS. We sample k uniformly spaced camera poses from an upper hemisphere around each scene. For adjacent image pairs, we compute the norm of the gradient predicted by a pre-trained Zero123~\cite{liu2023zero} model. The average gradient across all frames and scenes is defined as $\text{Zero123}_{grad}$.

For systematic evaluation, we used the first 64 prompts from Dreamfusion's~\cite{poole2022dreamfusion} online gallery of 397 prompts. The results, shown in Table.\ref{tab:zero123_grad}, indicate that our model achieves comparable $\text{Zero123}_{grad}$ to MVDream, while outperforming in R-Precision and A-LPIPS. This suggests our model maintains similar 3D awareness but offers superior texture quality.

We additionally quantified the occurrence of the Janus problem in our generated scenes. Instances with anomalies such as extra or missing parts — like an additional leg, a missing leg, or multiple heads — were identified as manifestations of the Janus problem. The results of this counting are detailed in Table. \ref{tab:janus_counting}.

\begin{table}[h]
\footnotesize  
\centering
\begin{tabular}{l c c c} 
\toprule
 Method & \# of Janus Scenes$\downarrow$ & \% of Janus Scenes $\downarrow$\\
\midrule
Dreamfusion\cite{poole2022dreamfusion} & 42 & 65.625\% \\
Prolificdreamer\cite{wang2023prolificdreamer} & 46 & 71.875\% \\
MVDream\cite{shi2023mvdream} & \textbf{3} & \textbf{4.688\%} \\
Ours & 5 & 7.813\% \\
\bottomrule
\end{tabular}
\caption{\textbf{Counting of Janus Scenes}}
\label{tab:janus_counting}
\end{table}

% \subsection{Generation Speed Comparison}
% \label{sec:speed}
\textbf{Speed Analysis:} Generation speed is a critical factor affecting usability. As illustrated in Table.~\ref{tab:speed}, we present the average time required to generate outputs for 64 prompts. Our method demonstrates a relatively faster generation speed compared to other methods, owing to our efficient decoupled optimization process.
% Since all these methods are optimization for a fixed timestep, the generation time didn't vary too much across prompts.

\begin{table}
\footnotesize  
\centering
\begin{tabular}{l c } 
\toprule
 Method & Time Per Scene $\downarrow$ \\
\midrule
Dreamfusion~\cite{poole2022dreamfusion} & 90mins  \\
Stable-Dreamfusion~\cite{stable-dreamfusion} w/ Instant-NGP~\cite{muller2022instant} & 11mins \\
Prolificdreamer\cite{wang2023prolificdreamer} & 210mins  \\
MVDream\cite{shi2023mvdream} & 90mins  \\
Our-SDS & 14 mins \\
Our-Decoupled & \textbf{5mins} \\
\bottomrule
\end{tabular}
\caption{\textbf{Generation Speed Analysis}}
\label{tab:speed}
\end{table}

%\subsection{Ablation Study}
%Our contribution includes orientation control diffusion model and decoupled back-propagation. 

%In Figure. 8, we show that while orientation control diffusion on its own addressing Janus problem  decoupled back-propagation greatly reduce noisy grain on the generated 3D assets. 

\subsection{Evaluation of Orientation Conditioned Image Generation.}
2D lifting-based text-to-3D methods are heavily dependent on the gradients generated by their 2D priors. Consequently, in order to understand the disparities among various text-to-3D methods, our initial focus is on the quantitative comparison and understanding of 2D diffusion models.

We employ three metrics: Fréchet inception distance (FID)~\cite{heusel2018gans}, Inception Score (IS)~\cite{IS}, and CLIP score~\cite{hessel2022clipscore}. FID and IS are used to measure image quality and CLIP score is used to measure text-image similarity. Results are presented in Table.~\ref{tab:2d_quantitative}. Scores are calculated over 397 text prompts from Dreamfusion's~\cite{poole2022dreamfusion} online gallery, and 10 images are generated for each prompt. The original StableDiffusion\cite{metzer2023latent} achieves the best score across three metrics. MVDream\cite{shi2023mvdream} get much worse FID after fine-tuning on Objaverse\cite{objaverse}. This partly explains why it experiences a reduction in texture quality and content diversity in generated 3D content. For comparison, our method achieves an FID of 16.93, much closer to the level of the original stable diffusion. This finding partially reveals why our model could produce photorealistic 3D models, and show the benefits of fine-tuning on MVImgNet\cite{yu2023mvimgnet}.

%We first use IS to compare the inherent quality of datasets.
%For ImageNet\cite{imgnet}, IS is calculated on 50k ImageNet validation set as reported in~\cite{barratt2018note}.
%For Laion, IS is calculated on 1000 LVIS\cite{gupta2019lvis} categories over 50k sampled images. For Objaverse, IS is calculated on 1230 categories over renderings of 3D assets\cite{shi2023mvdream}. For MVImgNet, IS is calculated on 

%Methods~\cite{liu2023zero, shi2023mvdream} fine-tuned on 3D dataset~\cite{objaverse} commonly exhibits the reduction of texture quality.
%As 3D asset dataset is usually manually crafted and have limited details compare to photo-realistic datasets like Laion-5B~\cite{schuhmann2022laion5b}, ImageNet~\cite{imgnet}, MVImgNet~\cite{yu2023mvimgnet}.

\begin{table}[t!]
\captionsetup{font=normal}
\setlength{\tabcolsep}{10pt}
\footnotesize
\begin{center}
\begin{tabularx}{1.0\linewidth}{X ccc}
\toprule
Model & FID$\downarrow$ & IS$\uparrow$ & CLIP$\uparrow$ \\
%\midrule 
%ImageNet\cite{imgnet}    & N/A & $63.70\pm 0.81$ & N/A \\
%Laion-5B\cite{schuhmann2022laion5b}   & N/A & 61.33 & N/A \\
%Objaverse\cite{objaverse} & N/A & $14.75\pm 0.81$ & $31.31\pm3.34$ \\
%MVImgNet\cite{yu2023mvimgnet} & N/A & N/A & N/A \\

\midrule

SD\cite{metzer2023latent}                 & \textbf{12.63} & $\textbf{15.75}\pm2.87$ & $\textbf{35.70}\pm1.90$ \\
MVDream\cite{shi2023mvdream}                 & 32.06 & $13.68\pm0.41$ & $31.31\pm3.12$ \\
\OURS    & 16.93 & $13.28\pm0.88$& $34.91\pm3.91$ \\
\bottomrule
\end{tabularx}
\caption{Quantitative evaluation on image synthesis quality.}
\label{tab:2d_quantitative}
\end{center}
\end{table}

%\subsection{Ablation Study}
%Our contribution includes orientation control diffusion model and decoupled back-propagation. 

%In Figure. 8, we show that while orientation control diffusion on its own addressing Janus problem  decoupled back-propagation greatly reduce noisy grain on the generated 3D assets. 
\section{Conclusion}

In summary, our proposed approach makes a modest yet meaningful contribution to the field of text-to-3D generation. The camera orientation conditioned diffusion model, a key aspect of our research, marks an advancement in generating 3D content with greater accuracy and consistency. We’ve circumvented significant challenges, such as the multi-head Janus issue, and refined the process of transforming text descriptions into 3D models. The introduction of the decoupled back propagation method represents our commitment to improving the efficiency and diversity of 3D model generation. We are hopeful that our work will provide a valuable foundation for future research in various applications of 3D technology.% and inspire further advancements 
%
%While we recognize that there is still much to explore and refine in this field, 
{
    \small
    \bibliographystyle{ieeenat_fullname}
    \bibliography{main}
}

% \input{section/X_suppl}

\end{document}